\documentclass{elsarticle}
\usepackage{hyperref}
\usepackage{float}
\usepackage{verbatim} %comments
\usepackage{apalike}
\restylefloat{figure}
\restylefloat{table}

\usepackage{booktabs}
\usepackage{verbatim}
\usepackage{makecell}
\usepackage{bm}
\usepackage{multirow}
\usepackage[]{graphicx}
\usepackage{todonotes}
\usepackage{multirow}
\usepackage{setspace}
\usepackage{amsmath}
\usepackage{comment}
\usepackage{booktabs}% http://ctan.org/pkg/booktabs

\usepackage{enumitem}
\usepackage{multirow}

\journal{Expert Systems with Applications}

\biboptions{authoryear}

\begin{document}
\begin{frontmatter}

\title{Summarization, Simplification, and Generation: \\ The Case of Patents}

\author[label1,label2]{Silvia Casola}
\ead{scasola@fbk.eu}

\author[label2]{Alberto Lavelli}
\ead{lavelli@fbk.eu}

\cortext[cor1]{Corresponding author.}
\address[label1]{Università di Padova, Human Inspired Technology Research Centre, Via Luzzatti, 4, 35121 Padova, Italy}
\address[label2]{Fondazione Bruno Kessler, Via Sommarive, 18, Trento, 38123, Italy}

\begin{abstract}
We survey Natural Language Processing (NLP) approaches to summarizing, simplifying, and generating patents' text. 
While solving these tasks has important practical applications -- given patents' centrality in the R\&D process -- patents' idiosyncrasies open peculiar challenges to the current NLP state of the art. 
This survey aims at a) describing patents' characteristics and the questions they raise to the current NLP systems, b) critically presenting previous work and its evolution, and c) drawing attention to directions of research in which further work is needed. 
To the best of our knowledge, this is the first survey of generative approaches in the patent domain. 
\end{abstract}


\end{frontmatter}

\section{Introduction}

Patents disclose what their creators consider valuable inventions -- so valuable, in fact, that they spend a nontrivial amount of time and money on protecting them legally. Not only do patents define the extent of the legal protection, but they also describe in detail the invention and its embodiments, its relation to prior art, and contain metadata.
It is common wisdom among patent professionals that up to 80\% of the information in patents cannot be found elsewhere~\citep{80tech}.

As a result, patents have been widely studied, with various aims. Recently, Natural Language Processing (NLP) approaches -- which aim at automatically analizyng text -- are emerging. 
This survey explores the application of NLP techniques to patent summarization, simplification, and generation. 
There are several reasons why we focus on these tasks: first of all, they have been explored less when compared, for example, to Patent Retrieval~~\citep{retrievalLupu, retreival2} and automatic patent classification~\citep{classification}. However, their practical importance is hard to overstate: the volume of patent applications is enormous (according to the World Intellectual Property Organization, WIPO, 3.2 million patents were filed in 2019), and keeping pace with the technology is becoming difficult. One of the  patents' aims is to make knowledge circulate and accelerate the transfer of technology: however, this is hardly achievable given the information overload.
Patent agents, R\&D groups, and professionals would thus highly benefit from tools that digest information from patents or make them easier to process, given their length and complexity.
The other reason is more technical and rises from patents' peculiar linguistic characteristics. Patents are primarily textual documents, and they have proved an interesting testbed for NLP researchers. Interesting yet challenging: being a mixture of legal and technical terms, patents' language differs severely from the general discourse. 

Our contributions are the following: we present an analysis of patents' linguistic characteristics and focus on the idiosyncrasies that negatively affect the use of off-the-shelf
NLP tools (Section \ref{section:primer}); after defining the patent summarization, simplification and generation tasks (Section \ref{section:tasks}) we describe the few available datasets and the evaluation approaches (Sections \ref{section:data} and \ref{section:eval}). Next, we review previous work in Sections \ref{section:summarization}, \ref{sec:simplification}, and \ref{sec:generation}. Our review is rather comprehensive and covers works from the early 2000s to date. We pay special attention to the algorithms and models used from an NLP perspective. To the best of our knowledge, this is the first work that surveys summarization, simplification and generation techniques specifically in the patent domain. Note that, however, since patent processing has historically been application-oriented, previous work often used project-specific datasets, making it difficult to compare approaches directly in terms of performance. 
Finally, we present interesting lines of investigation for future research.

\section{A primer on patents}
\label{section:primer}
Patents are primarily legal documents. Their owner controls the use of an invention for a limited time in a given geographic area and thus excludes others from making, using, or selling it without previous authorization. In exchange, the inventor discloses the invention to facilitate the transfer of technology.

This section defines some domain-specific concepts that we will reference in the following; we use patent US4575330A\footnote{\url{https://patents.google.com/patent/US4575330A/en} [Last accessed: March 2021] } (the antecedent of a 3D printer, designed by Hull in 1989) as a running example.

\subsection{Patent documents}
Patent documents are highly structured and must follow strict rules~\footnote{WIPO Patent Drafting Manual (2007). URL: \url{https://www.wipo.int/publications/en/details.jsp?id=297} [Last accessed: May 2020].}.
Typically, they contain the following sections: 
\begin{description}
\item [Title] E.g.,  \textit{Apparatus for production of three-dimensional objects by stereolithography}

\item [Claim] Specifies the extent of legal protection. This section can include multiple claims\footnote{We will refer to the whole document section using the cased form \textit {Claim}, while the individual \textit{claims} contained in such section will be lowercase.} with a hierarchical structure.
\textit{
\begin{enumerate}
    \item  A system for producing a three-dimensional object from a fluid medium capable of solidification when subjected to prescribed synergistic stimulation, said system comprising: means for drawing upon and forming successive cross-sectional laminae of said object at a two-dimensional interface; and means for moving said cross-sections as they are formed and building up said object in step wise fashion, whereby a three-dimensional object is extracted from a substantially two-dimensional surface.
    \item An improved system for producing a three-dimensional object from a fluid medium capable of solidification when subjected to prescribed synergistic stimulation, said system comprising: [...]
    \item A system as set forth in claim 2, and further including: programmed control means for varying the graphic pattern of said reaction means operating upon said designated surface of said fluid medium.
\end{enumerate}
}

Claims 1 and 2 are independent, while claim 3 is dependent on claim 2, which it further specifies. The document comprises 47 claims, which this paper is too small to contain. Following patent rules, each claim consists of a single sentence, therefore long, complex, and highly punctuated. The language is abstract to obfuscate the invention's limitations and full in legal jargon.

\item [Description] A description detailed enough for a person skilled in the art\footnote{A ``person skilled in the art'' has ordinary skills in the invention technical field. For a formal definition, refer to the \hyperlink{https://www.wipo.int/export/sites/www/pct/en/texts/pdf/ispe.pdf}{PCT International Search and Preliminary Examination Guidelines}.} to make and understand the invention.
\bigbreak
\hfill\begin{minipage}{\dimexpr\textwidth-1.7cm}
\textit{ Briefly, and in general terms, the present invention provides a new and improved system for generating a three-dimensional object by forming successive, adjacent, cross-sectional laminae of that object at the surface of a fluid medium capable of altering its physical state in response to appropriate synergistic stimulation, the successive laminae being automatically integrated as they are formed to define the desired three-dimensional object.
}
\end{minipage}
\smallbreak
\hfill\begin{minipage}{\dimexpr\textwidth-1.7cm}
\textit{In a presently preferred embodiment, by way of example and not necessarily by way of limitation, the present invention harnesses the principles of computer generated graphics in combination with stereolithography, i.e., the application of lithographic techniques to the production of three dimensional objects, to simultaneously execute computer aided design (CAD) and computer aided manufacturing (CAM) in producing three-dimensional objects directly from computer instructions. [...]
}
\end{minipage}
\newline

While the Claim section aims at legally protecting the invention (the construct in the mind of the inventor, with no physical substance), the Description discloses one or more embodiments (physical items). Drawings are standard in this section. The Description illustrates the invention to the public on the one hand and supports the Claim on the other. Notice how, while the language is still convoluted, it is less abstract. 
\item [Abstract] Summarizes the invention description. 
\bigbreak
\hfill\begin{minipage}{\dimexpr\textwidth-1.7cm}
\textit{
A system for generating three-dimensional objects by creating a cross-sectional pattern of the object to be formed at a selected surface of a fluid medium capable of altering its physical state in response to appropriate synergistic stimulation by impinging radiation, particle bombardment or chemical reaction, successive adjacent laminae [...].
}
\end{minipage}

\item [Other metadata] Includes standard classification codes, prior art citations, relevant dates, and inventors', assignees', and examiners' information.   
\item[Patent classifications]
Patents are classified using standard codes. The Patent Classification (IPC)\footnote{\url{wipo.int/classifications/ipc/en/} [Last accessed: March 2021]} and the Cooperative Patent Classification (CPC)\footnote{\url{cooperativepatentclassification.org} [Last accessed: March 2021]} are the most widespread. Patent examiners assign codes manually depending on the invention's technical characteristics.
Patent US4575330A has 14 IPC classification codes. For example, code G09B25/02 indicates that the patent is in the Physics (G) section and follows to specify the class (G09), sub-class (G09B), group (G09B25/00), and sub-group (G09B25/02). 
\end{description}

\subsection{Patent language}
In this section, we describe what makes patent documents unique from a linguistic perspective. 
Few documents are, in fact, as hard to process (for both humans and automatic systems) as patents, with their obscure language and complex discourse structure.  
\begin{description}
\item[Long sentences] According to patents' rules, each claim must be written in a single sentence, which is therefore particularly long. \cite{nlp_challenges} examined over 67 thousand Claim sections and found a median length of 22 and a mean of 55; note that this figure is highly underestimated, as the authors segment sentences using semicolons in addition to full stops. In contrast, they found that the British National Corpus median length (when segmented using the same methodology) is less than 10.
For comparison, the first claim in patent US4575330A (a ``rather short'' one) is 69 words long, while claim 2 contains 152 words.
%; the BNC contains samples from several sources (news, novels, letters, essays) and is thus used as a standard for English.
\cite{jap} found similar characteristics in Japanese. While most quantitative work focuses on the Claim, sentences in other sections are also remarkably long.     

\item[Words' distribution and vocabulary]Claims do not use much lexicon not covered in general English, but their word frequency is different, and novel technical multi-word terms are created \textit{ad hoc}~\citep{nlp_challenges}. Moreover, many words are used unusually: \textit{said}, for example, typically refers back to a previously mentioned entity, repeated to minimize ambiguity (e.g., \textit{A system for [...], said system comprising [...]}, in claim 1); transitions (e.g.,  \textit{comprising}, \textit{including}, \textit{wherein}, \textit{consisting}) have specific legal meanings. 
The Claim's language is abstract (\textit{system}, \textit{object}, \textit{medium} in claim 1), not to limit the invention's scope, while the Description is more concrete~\citep{genrespecific}.

\item[Complex syntactic structure] Patent claims are built out of noun phrases instead of clauses, making it nontrivial to use general NLP resources. As a result, previous work has tried to adapt existing parsers with domain-specific rules~\citep{parser1} or simplify the claim before parsing~\citep{multilingual}.
\end{description}

\section{Task description}
\label{section:tasks}
In this section, we will discuss the tasks of text summarization, simplification, and generation. We will define them from an NLP perspective and discuss their practical importance in the patent domain. 


\subsection{Summarization}
Loosely speaking, a summary is a piece of text that, based on one or more source documents, 1) contains the main information in such document(s) and 2) is shorter, denser, and less redundant. 
For a recent survey on text summarization, see~\citep{surveysumm}.
Automatic summarization is an open problem in modern Natural Language Processing, and approaches vary widely.
We will categorize previous work according to the following dimensions: 
\begin{description}
    \item [Extractive vs. abstractive] Extractive summaries consist of sentences or chunks from the original document. To this end, most approaches divide the input into sentences and score their relevance. 
    %Thus,  the top sentences should carry the document's key content. 
    In contrast, abstractive approaches build an intermediate representation of the document first, from which they generate text that does not quote the input verbatim. Finally, hybrid systems take from both approaches; for example, they might select sentences extractively and then generate a paraphrased summary. Patent summaries have traditionally been extractive, but an interest in abstractive summarization is emerging.    
    \item [Generic vs. query-based] %Most summarization approaches generate a summary that distills the source without further constraints.
    Query-based models~\citep{query1, query2, query3} receive a query and summarize information of relevance to such query. For example, during a prior art search, the user might only be interested in aspects of the retrieved documents that might invalidate their patent.
    % \item[Single- vs. multi-document input] Single-document systems summarize a single input, while some systems can process a varying number of documents. Given patents' length, summarizing multiple patents is difficult and has scarcely been explored.
    \item[Human- vs. machine-focused] While summaries are typically intended for humans, producing a shorter dense representation is equally relevant when the input is too long to be processed directly,  e.g., by a machine learning algorithm. In this case, summarization constitutes a building block of a more complex pipeline. \cite{TSENG1, TSENG2}, for example, perform summarization in view of patent-map creation and classification.     
    %\item[Claim-based vs. Description-based] Given the length of patent documents, some previous work has focused on specific sections: either the Claim - for its legal importance - or the Description, given its more concrete language. Some approaches use both sections, given their complementary nature~\citep{genrespecific}.
    \item[Language-specific vs. multilingual] While published research has primarely been anglocentric, some works in other languages and multilingual techniques have been proposed. 
\end{description}

As expected, patents' summarization comes with its challenges. For example, while in some domains (e.g., news) the essential facts are typically in the first paragraphs, this assumption does not hold for patents, whose important content is spread in the whole input. Summaries also contain a high percentage of n-grams not in the source and shorter extractive fragments. Finally, summaries' discourse structure is complex, and entities recur in multiple sentences. All these characteristics make patents an interesting testbed for summarization, for which a real semantic understanding of the input is crucial~\citep{bigpatent}.

In addition to the research interest, patents summaries are practically relevant for R\&D teams, companies, and stakeholders. A brief search of online services showed that some companies sell patent summaries and related data as a paid service. For example, Derwent\footnote{\url{https://clarivate.com/derwent} [Last accessed: March 2021]} produces patent abstracts distilling the novelty, use and advantages of the invention in plain English; to the best of our knowledge, the abstract is manually compiled by experts.

\subsection{Simplification}
Automatic simplification reduces the linguistic complexity of a document to make it easier to understand.  In contrast with summarization, all information is usually kept in the simplified text.   
Generally, approaches vary depending on the system's target user (e.g., second-language learners, people with reading disorders, children).~\citet{simplification} is a recent survey addressing text simplification in the general domain.
Given patents' complexity -- lexically and syntactically -- the challenge lies in making their content accessible to the lay reader (which justifiably gets scared away from patents) and simplifying the experts' work.  

We will consider the following aspects:
\begin{description}
\item[Expert vs. lay target reader] Patents' audience ranges from specialists (e.g., attorneys and legal professionals), to laypeople (including academics) that might be interested, for example, in the invention's technical features. Depending on the target user (and, in turn, on the target task), the degree of simplification might vary. When considering the legal nature of patents, for example, special attention should be given to keeping their scope unchanged. The first claim of patent US4575330A, for example, states: \textit{``A system for producing [...] comprising: means for drawing [...]; and means for moving [...].''}.
A system ``comprising'' a feature might include additional ones; thus,  replacing the term with ``consisting of'' -- which, in patent jargon, excludes any additional component -- would be problematic, even if thesauruses treat the terms as synonyms\footnote{see, for example, \hyperlink{https://www.epo.org/law-practice/legal-texts/html/guidelines/e/f_iv_4_20.htm}{Collins Online Thesaurus}.}. Obviously, the attention to the jargon can be loosened if the target user is more interested in the technical characteristics than in the legal scope.

\item[Textual vs. graphical output] 
The simplification system's output can be either a text or a more complex data structure. A textual output can be formatted appropriately (e.g., coloring essential words~\citep{informationExtraction}), annotated with explanations (e.g., with links from a claim to a Description passage~\citep{allignJap}), or paraphrased~\citep{improving}. 
Alternatively, a graphical representation, in the form of trees or graphs -- which e.g.  highlights the relation among the invention components -- can be used. 

\item[Application] The simplification system can be designed with a specific application in mind: in~\citep{informationExtraction}, for example, authors designed an interface to help patent experts in comparing documents from the same patent family. 
\end{description}

As in the case of summaries, designing appropriate simplification systems has interesting use cases. \cite{evalSim} performed a user study with both experts and laypeople: most of their participants considered patents difficult to read. When presented with various reading aids, most considered them useful. Even law scholars have called for the use of a simpler language in patents~\citep{plain}.
Commercially, companies that provide patent reports do so in plain language. Somewhat ironically, Derwent goes as far as replacing the document title with a less obscure one, of more practical use. 

\subsection{Generation}
%Natural Language Generation is an NLP branch that aims to generate new, original text automatically. %This definition includes summarization and simplification as text-to-text instances. 
We will use Patent Generation to refer to methods that aim at generating a patent or part of it. 
To the best of our knowledge, this line of research is relatively new and is likely inspired by the recent success of modern generative models (e.g. GPT and its evolutions ~\citep{gpt, gpt2, gpt3}) in various domains, including law~\citep{law}, health~\citep{health} and journalism~\citep{journalism}, to name a few.

Some approaches only produce ``patent-like'' text (i.e., employing technical terminology and respecting patents' writing rules): their generation is unconstrained or constrained to a short user prompt -- the first words of a text that the system needs to extend coherently.
Their practical use is likely limited, but their success shows that even patents' obscure language can be mastered by machines, at least at a superficial level. Another class of approaches conditions the generation to a fragment of the patent to produce a coherent output.
For example, one might want to produce a plausible patent Abstract given its Title or a set of coherent claims with a given Description. In this case, the generation is constrained to the whole input section (e.g, the Title text) and the type of output section (e.g., Abstract).

While patent generation is still in its early days, researchers dream of ``augmented inventing''~\citep{gpt2pat}, assisting inventors in redefining their ideas and helping with patent drafting. To this end, some hybrid commercial solutions are already in the market\footnote{see, for example \url{https://bohemian.ai/case-studies/automated-patent-drafting/},  \url{https://www.patentclaimmaster.com/automation.html}, \url{https://harrityllp.com/services/patent-automation/} [Last accessed: March 2021]}.

\section{Datasets}
\label{section:data}
Patent documents are issued periodically by the responsible patent offices. The United States Patent and Trademark Office (USPTO), for example, publishes patent applications and grants weekly, along with other bibliographic and legal data\footnote{\url{developer.uspto.gov/data} [Last accessed: March 2021]}. To access the documents programmatically, Application Programming Interfaces (APIs) are available. PatentsView\footnote{\url{www.patentsview.org/} [Last accessed: March 2021]}, for example, is a visualization and mining platform to search and download USPTO patents, updated every three months. It provides several endpoints (patent, inventor, assignees, location, CPC, etc.) and a custom query language. 
Google also provides public datasets\footnote{\url{console.cloud.google.com/marketplace/browse?q=google\%20patents\%20public\%20datasets\&filter=solution-type:dataset} [Last accessed: March 2021]}, accessible through BigQuery.

While it is relatively easy to obtain raw patent text, few cured datasets exist. These data are of the greatest importance: having a set of shared benchmarks allows to directly compare approaches, which is much more difficult otherwise.
The only large-scale dataset for patent summarization is BigPatent\footnote{\url{evasharma.github.io/bigpatent} [Last accessed: March 2021]}~\citep{bigpatent}. The dataset was recently built for abstractive summarization and contains 1.3 million patents' Descriptions and their Abstracts (a.k.a. Summary of Description) as human-written references. While most previous work focuses on Claims' summarization, no comparable Claim to summary dataset exists (nor would it be easy to obtain), and authors resort to expert-written summaries for evaluation.

For patent simplification, no simplified corpus exists to date.

\section{Evaluation}
\label{section:eval}
The evaluation of a generated text, be it a summary, a simplification, or a completely new document, is currently an open problem in Natural Language Generation~\citep{eval, evalSurvey2}. 
Qualitative approaches resort to humans to evaluate the generated text (either overall or in some specific dimensions, e.g., relevancy, coherence, readability, redundancy) and are to date considered the gold-standard for evaluation. In contrast, automatic approaches usually measure the output similarity with human written gold-standards (e.g. ROUGE \citep{rouge}, BLUE \citep{bleu}, and PYRAMID \citep{pyramid}); while not perfect, automatic metrics have a certain degree of correlation with human judgment and are used when performing human evaluation is too expensive or labour-intensive. 

For patent summarization, qualitative evaluation involves experts and non-experts; \cite{multilingual}, for example, assess summaries intelligibility, simplicity, and accuracy on a Likert scale \citep{likert}. 
Quantitatively, the most widespread automatic summarization metrics is ROUGE (Recall-Oriented Understudy for Gisting Evaluation)~\citep{rouge}. It measures the overlap between the generated sentence and the gold-standard. ROUGE-N is n-gram based and is measured as: 
$$\begin{aligned} ROUGE{-}N = \frac{\sum _{S \in Reference} \sum _{gram_{n} \in S}  Count_{match}(gram_{n})}{\sum _{S \in Reference} \sum _{gram_{n} \in S}  Count(gram_{n})}. \end{aligned}$$
ROUGE-L measures the similarity in terms of the Longest Common Subsequence (LCS). Words of the LCS must appear in the same relative order but not necessarily be contiguous. %Given a candidate summary of length $l$ and a reference of length $m$, the precision and recall metrics are $P=\frac{LCS(Generated, Reference)}{l}$ and $R=\frac{LCS(Generated, Reference)}{m}$ and the final score is their weighted harmonic mean.  
ROUGE-1, ROUGE-2 (for relevance), and ROUGE-L (for fluency) are generally used in practice, as they best correlate with human judgment.
Similarly, some studies measure the similarity between the generated text and the reference summary in uni-gram Precision, Recall, and $F_1$.
The Compression Ratio and the Retention Ratio (the percentage of original information kept in the summary) are also frequently reported.  
Finally, when summarization is part of a more complex pipeline, the relative improvement of the downstream task is considered.  

When evaluating simplification approaches, two different points of view exist. The first only considers the method's correctness: if the algorithm needs to segment the text, one can manually annotate a segmented gold-standard and measure accuracy. However, assessing the readability improvement requires qualitative studies. \cite{evalSim}, for example, use a questionnaire for quantifying patents'  complexity and test simplification solutions.  
Following their work's findings, experts' and laypeople's opinions should be analyzed separately, as they are concerned with different issues. For instance, experts worry that the simplified patent might be misrepresented and its legal scope changed while laypeople demand strategies to understand the invention and find information.    

Finally, measuring the quality of generated patent text is generally tricky. When no gold-standard exists, some authors have introduced \textit{ad hoc} measures (see, for example~\citep{measuringSpan}); when a human-written reference exists, metrics as ROUGE can be used.
Finally, note that some studies criticize the use of ROUGE; ~\cite{metadata}, for example, also reports the results using the Universal Sentence Encoder~\citep{use} representation, which they speculate handles semantics better. 

\section{Approaches for patent summarization}
\label{section:summarization}
In this section, we describe extractive and abstractive approaches to patent summarization. As we discussed already, their direct comparison is difficult, as publications tend to use slightly different tasks on unshared data. 
The approaches discussed in the paper are summarized in Table \ref{tab:summarization}.

\begin{table}[]
\resizebox{\textwidth}{!}{%
\begin{tabular}{@{}llllll@{}}
\toprule
Study & Approach & Main contribution & Limitations & Dataset \\ 
\midrule
\makecell[l]{\parbox[c]{2cm}{\cite{TSENG1, TSENG2}\\}} & Extractive & \makecell[l]{Domain-specific \\ considerations;\\key-phrase extraction \\algorithm} & \makecell[l]{Extrinsic eval. \\only (classification \\ surrogates)} & \makecell[l]{National Science\\Council Patent\\Set (612 patents)}\\
& \\
\makecell[l]{\parbox[c]{2cm}{\cite{trappey1,trappey2}}} & \makecell[l]{Extract\\information-dense\\ paragraphs} & \makecell[l]{Application of \\general-domain\\ techniques} & Evaluation & \makecell[l]{111 patents}\\
& \\
\parbox[c]{2cm}{\parbox[c]{2cm}{\cite{trappeysemantic, trappeysemantic2}}} & \makecell[l]{Extract\\ information-dense\\ paragraphs} & \makecell[l]{Ontology for \\key-phrase\\extraction}& & 200 patents \\
& \\
\parbox[c]{2cm}{\cite{multilingual}} & \makecell[l]{Abstractive\\(Deep-Syntactic\\Structs)} & Multilinguality & Complexity & 50 patents \\
&\\
\parbox[c]{2cm}{\cite{topas,genrespecific}} & Hybrid & \makecell[l]{Patent-specific\\approach (lexical\\chain, Claim-\\Description alignment,\\sentence fragmentation)} & Complexity & \makecell[l]{26 patents\\(test)}\\
&\\
\makecell[l]{\parbox[c]{2cm}{\cite{query3, query1, query2}}} & \makecell[l]{Extractive\\(query-oriented)} & \makecell[l]{Query-oriented\\approach\\Query expansion\\strategies} & & \makecell[l]{Smartphone\\-related\\ patents} \\
& \\
& \\
\parbox[c]{2cm}{\cite{bigpatent}} &  & Dataset & \makecell[l]{ Complex Abstract\\ style} & 1.3M patents \\
&\\
\makecell[l]{\parbox[c]{2cm}{\cite{lsa}}} & \makecell[l]{Extractive,\\semantic similarity} & \makecell[l]{Summarization to\\name patent\\ groups} & & \makecell[l]{733 patents\\(test)}\\
&\\
\parbox[c]{2cm}{\cite{trappey2020}} & \makecell[l]{Hybrid (abstractive\\to extractive)} & \makecell[l]{Attention-based\\method for\\ extracting\\keywords} & \makecell[l]{Complexity} & \makecell[l]{1708 (train)\\ 30 (test)\\patents}\\
&\\
\parbox[c]{2cm}{\cite{pegasus}} & \multirow{3}{*}{\makecell[l]{Abstractive\\(transformer-based)}} & \multirow{3}{*}{\makecell[l]{Analysis of SOTA \\ general-domain NLP \\systems in the patent\\domain}} & \multirow{3}{*}{\makecell[l]{Data requirements\\Computational\\cost}} & \multirow{3}{*}{BigPatent} \\
\parbox[c]{2cm}{\cite{ctrl}} &  & \\
\parbox[c]{2cm}{\cite{bigbird}} & & \\
&\\
\parbox[c]{2cm}{\cite{namingabstractive}} & \makecell[l]{Abstractive (LSTM),\\semantic similarity} & \makecell[l]{Summarization to name\\patents group} & \makecell[l]{Abstractive\\approaches\\inferior to\\extractive\\ones} & \makecell[l]{ 41,527
(train), \\733 patents\\(test)} 
\\
\\ \bottomrule
\end{tabular}
}
\caption{Surveyed studies for Patent Summarization.}
\label{tab:summarization}
\end{table}

\subsection{Extractive summarization}
Extractive approaches select the most informative sentences in the original document. 
A typical pipeline comprises the following steps:
\begin{enumerate}
    \item Document segmentation: documents are split into segments, sentences, or paragraphs, using punctuation or heuristics. While many approaches work at the sentence level, \cite{genrespecific} argue that patents sentences are too long to be used directly, and further segment them. In many cases, only some Sections (e.g. Description, Claims) are considered. 
    \item Sentence preprocessing: includes standard text preprocessing, e.g., removing stopwords or stemming. Given the peculiar patent style, patent-specific stopwords (cured by experts) also need to be removed. Some approaches~\citep{trappey1,trappeysemantic} only keep specific Parts of Speech.  
    \item Feature extraction: for each sentence, general-domain features include keywords, title words, cue words (from expert-designed lists), and the sentence position. In particular, patents contain several multi-word entities that need to be identified. To this end,~\cite{TSENG1} propose an algorithm that merges nearby uni-grams words and extracts maximally repeated strings as multi-word terms.  Given that text is often full in technical terms, \cite{trappeysemantic, trappeysemantic2} use a domain ontology for identifying domain-specific key-phrases. 
    The approaches above try to customize general-discourse features to the patent domain; in contrast, \cite{genrespecific} propose a domain-specific approach. They consider the lexical chain length as a measure of an entity importance: i.e., invention components that appear many time in the Claim and Description are particularly relevant. Given the abnormal patents' sentences length, they further segment sentences and use fragments as extractive candidates. 
    
    \noindent
    In most approaches, the segment position is also be considered (favoring sentences at the beginning of a paragraphs or paragraphs at the beginning or end of a Section). 
    Query-oriented approaches also measure the sentence similarity to the query (e.g., with overlapping words~\citep{query3}), which can be further expanded using a domain ontology~\citep{query2} or general-domain resources~\citep{query1} like WordNet. Query expansion can be particularly important as different patent documents can purposely use a completely different wording for similar components. 
    Table \ref{table:extractive} includes some frequent features in extractive patent summarization. 
    \item Sentence weighting: the extracted features are used to score the sentence relevance in the summary. For example, \cite{TSENG1} score sentences as:
    $$ score(S)=\Bigg(\sum_{w \in {key_w, title_w}} {TF_w} + \sum_{w \in clue_w} mean(TF)\Bigg) \times FS \times P $$
    where TF is the term frequency of word w in sentence S, mean(TF) is the average term frequency over keywords and title words in S, and FS and P are the sentence position weights, assigned heuristically. In particular, FS is set to 1.5 if the sentence is the first in the paragraph and to 1 otherwise; P is the position weight of the sentence with respect to the Section, and is set to 2 or 4 if the sentence is in the first or last two paragraphs of the Section respectively, and to 1 otherwise.  
    
    \noindent
    Another option is to learn weights from data directly: for example,~\cite{genrespecific} score each segment as $score(S) = \sum^{n}_i w_i f_i$; they use linear regression to learn features weights based on textual segments and their cosine similarity to the gold-standard. Lastly, sentences can be classified as relevant or not relevant: to this end, \cite{query2, query3} train a Restricted Boltzmann Machine~\citep{rbm} without supervision.  
    To minimize repetitions, \cite{trappey1,  trappeysemantic, trappey2} cluster semantically similar sentences and only select one sentence per cluster. 
    \item Summary generation: most commonly, the final summary consists of the union of the extracted sentences. \cite{trappeysemantic, trappeysemantic2} also draw a summary tree linked to the domain ontology. 
\end{enumerate}

While popular, the above pipeline is not the only route to extractive summarization. 
Alternatively, ~\cite{improving} exploit patent's complex discourse structure, which they prune following predefined domain-specific rules. Finally,~\cite{lsa} discuss applying general-domain algorithms to 
%The former is a graph-based method. The algorithm computes the cosine similarity among sentences based on their bag-of-words representation and store them in a matrix. The similarity matrix is then interpreted as an adjacent matrix for a graph - having sentences as nodes and their similarity as edges - and a threshold is set to remove weak edges. The most central sentences (i.e., nodes having the highest degree) are extracted.
%The latter method defines a sentence-term matrix, which it decomposes using Singular Value Decomposition (SVD). The sentences associated with the biggest eigenvalues are extracted.
patent sub-groups naming\footnote{Patent sub-groups are the most specific level of the patents' classification hierarchy and are named with a representative name, e.g. ``Extracting optical codes from image or text carrying said optical code''.}: in that context,  LSA~\citep{lsasumm} performs best compared to LexRank~\citep{lexrank} and to a \textit{TF-IDF} approach.

\begin{table}[]
\begin{center}
\resizebox{\textwidth}{!}{%
\begin{tabular}{@{}cc@{}}
 Features & Description \\
 \hline
 \multicolumn{2}{l}{\textbf{Entity features}} \\
Term frequency - Inverse Document Frequency & Measures a keyword importance \\
Ontology-based & \makecell{Concepts from a domain-specific ontology; \\ specific concepts are more relevant} \\
Coreference-chain based & Entities coreferenced repeatedly are more central \\
& \\
 \multicolumn{2}{l}{\textbf{Segment features}} \\
 Title similarity & \multirow{3}{*}{\makecell{Computed by considering either word overlap \\ or semantic similarities}} \\
 Abstract similarity \\
 Claim similarity \\
 Query similarity & Relevance to the query   \\
 Position & \makecell{Patent section (Claim, Description, etc) \\ and sentence position within the section} \\
 Length & Overly long segments might be discouraged \\
 Number of keywords & \\
 Number of cue-words & \\
 
\hline
\end{tabular}
}
\end{center}
\caption{Extractive features. We use the term \textit{entity} to generically refer to keywords, phrases or other mentions in the document. Similarly, \textit{segment} indicated both complete sentences and fragments.}
\label{table:extractive}
\end{table}
 
\subsection{Abstractive models}
Abstractive  models exploit a semantic representation of the input.
In the patent domain, the first approaches used deep syntactic structures. %In the context of the PATExpert project~\citep{patexpert},
Given patents' linguistic structure, \cite{multilingual} need to first simplify the claims (see~\citep{simplificationShallow}) to achieve adequate parsing performance; then, they map the shallow syntactic structures to deep ones, using rules. Deep syntactic structures are closer to a semantic representation and thus used for summarization: to this end, the least relevant chunks are removed using handcrafted rules. Finally, they transfer the summarized deep structures to the target language (English, French, Spanish, or German) and use a generator to convert them to text. 

More recently, neural models have revolutionized Natural Language Processing. These models act on the text directly and use neural networks to extract a representation optimized for the task to be solved. For abstractive summarization, a sequence-to-sequence model typically extracts a hidden representation from the input text (encoding) and then uses it to generate the output (decoding). While neural performance is indisputable, models require many input-output samples to learn from: that is probably why they have only spread very recently in the patent domain. No large-scale summarization dataset, in fact, existed before 2019, when BigPatent~\citep{bigpatent} was published. 
Sharma et al. proposed several baselines: an LSTM~\citep{seq2seq} with attention \citep{attention}, a Pointer-Generator~\citep{pointer} with and without coverage, and SentRewriting~\citep{sentrewriting} (a hybrid approach). 

Given its differences with the previously available datasets (mostly in the news domain) -- in terms of style, content distribution and discourse structure --, BigPatent became an interesting testbed for general domain NLP summarization models: this is the case of  Pegasus~\citep{pegasus}, a pre-trained transformer~\citep{transformer} for summarization. During pre-training, whole sentences from the input are masked, and the model needs to generate them from the rest of the input (Gap Sentence Generation).  

One of the significant challenges of the dataset is the input length, which is very large (with a 90\% percentile of 7,693 tokens), and is problematic for standard transformers (whose attention mechanism scales quadratically in the input size): to this end, BIGBIRD~\citep{bigbird} proposes a sparse attention mechanism which, to the best of our knowledge, is  to date the state of the art on the dataset.

\begin{table}[]
\begin{center}
\begin{tabular}{lccc}
Model      & R-1   & R-2   & R-L \\
\hline
TextRank~\citep{textrank}   & 35.99 & 11.14 & 29.60 \\
LexRank~\citep{lexrank}    & 35.57 & 10.47 & 29.03 \\
SumBasic~\citep{sumbasic}   & 27.44 & 7.08  & 23.66 \\
RNN-ext RL~\citep{sentrewriting} & 34.63 & 10.62 & 29.43 \\
\hline
LSTM seq2seq~\citep{seq2seq} + attention   & 28.74 & 7.87  & 24.66 \\
Pointer-Generator~\citep{pointer} & 30.59 & 10.01 & 25.65 \\
Pointer-Generator + coverage~\citep{pointer} & 33.14 & 11.63 & 28.55 \\
SentRewriting~\citep{sentrewriting} & 37.12 & 11.87 & 32.45 \\
\hline
TLM~\citep{tlm} & 36.41 & 11.38 & 30.88 \\
TLM + Extracted sentences & 38.65 & 12.31 & 34.09 \\
\hline
CTRL$_{sum}$~\citep{ctrl} & 45.80 & 18.68 & 39.06 \\
\hline
Pegasus$_{base}$~\citep{pegasus} (no pretraining) & 42.98 & 20.51 & 31.87 \\
Pegasus$_{base}$  & 43.55 & 20.43 & 31.80 \\
Pegasus$_{large}$  (C4) &  53.63 & 33.16 & 42.25 \\
Pegasus$_{large}$ (HugeNews) & 53.41 & 32.89 & 42.07 \\
\hline
\makecell[l]{BIGBIRD-RoBERTa\\\hspace{1em}(base, MLM)~\citep{bigbird}} & 55.69 & 37.27 & 45.56 \\
BIGBIRD-Pegasus (large, Pegasus pretrain) &
\textbf{60.64} & \textbf{42.46} & \textbf{50.01}

\end{tabular}
\end{center}
\caption{Results on the BigPatent dataset. TextRank, LexRank, SumBasic, and RNN-ext RL are extractive baselines. TLM uses a GPT-like transformer (TLM) and concatenates extracted sentences to the Description (TLM + Extracted sentences). Results reported for CTR refer to unconditioned summarization. For Pegasus, we report results for base model (223M parameters) with and without pre-training and a larger model (568M parameters) independently pre-trained on a dataset of web pages (C4) and a dataset of news articles (HugeNews). For BIGBIRD, results using RoBERTa's (MLM) and a Pegasus' (Gap Sentence Generation) pre-training are considered. }
\label{tab:bigpatent}
\end{table}

Summarization models' performance on the BigPatent dataset is shown in Table \ref{tab:bigpatent}. Note how the pre-trained transformer models obtain the best results, in line with the general trend in Natural Language Processing.

Finally, summarization methods could also be used for solving specific patent tasks. 
CTRLsum~\citep{ctrl}, for example, is a system that allows controlling the generated text by interacting through keywords or short prompts. The authors experiment with inputting \textit{[the purpose of the present invention is]} to retrieve the patent aim.
Finally,~\cite{namingabstractive} have compared extractive and abstractive models in naming patents' subgroups. When used to ``summarize'' the Abstract to produce a patent Title -- which should contain, similarly to its subgroup name, the essence of the invention -- extractive methods were found superior. This result highlights the challenges met by abstractive models, which are likely to be magnified in the legal domain. 

\subsection{Hybrid models}
Hybrid models integrate elements of extractive and abstractive summarization. 
%For example, the TOPAS workbench~\citep{topas} includes a module that first selects segments extractively and then aggregates them using deep linguistic structures.  
%In addition to typical extractive features (TF-IDF, position, Abstract, Title and Claim similarity of the sentence), it uses domain-specific features, both entity-related (frequency, coreference chain, if an entity appears in the Claim, etc.) and segment-related (similarity to the Claim, length, etc). Features weights are learned using linear regression.
%paraphrased and turned into grammatical text. Similarly,~\citep{genrespecific} uses entity distribution and lexical-chain features to extract relevant Claim and Description segments.
For example, the TOPAS workbench~\citep{topas} includes a module that first selects segments extractively and then paraphrases them. A similar approach was adopted in~\citep{genrespecific}. In this approaches, a sentence fragment is the unit of extraction (sentences are too long to be used directly); extracted fragments are then paraphrased. %Similarly,~\citep{genrespecific} uses entity distribution and lexical-chain features to extract relevant Claim and Description segments.
More recently,~\cite{tlm} have shown that adding previously extracted sentences to the input when training a language model helps with long dependencies and improves the model's abstractiveness. 
While the models described so far train the extractive and the abstractive components separately, SentRewriting~\citep{sentrewriting} uses reinforcement learning for selecting salient sentences and train the model end to end. The last two mentioned models are general-domain, and also test their results on patents.

In contrast with the previous works, ~\cite{trappey2020} explore an abstractive to extractive approach. They use an LSTM with attention to guide the extraction of relevant sentences: it receives a set of English and Chinese documents (Title, Abstract, and Claim) and is trained to produce a human-written summary (abstractive component). After the training, the words with the highest attention weights are retrieved and treated as automatically-extracted keywords; sentences are then scored and extracted accordingly (extractive component). This approach is domain-specific, and is used as a way to simplify the keyword extraction, which is complex in the patent domain. 
 
\section{Approaches for Patent simplification}
\label{sec:simplification}
Patents' claims are the hardest section of an overall hard-to-read document. As such, a lot of effort has been spent in improving the accessibility and readability of the Claim. Table \ref{tab:simplification} summarized previous work.

Given the Claim's legal nature, however, the extent of the modification is crucial, and previous approaches' views to the task have varied widely. 
\begin{table}[]
\resizebox{\textwidth}{!}{%
\begin{tabular}{@{}lllll@{}}
\toprule
Study & Approach & Main contribution & Limitations & Dataset \\ \midrule
\makecell[l]{\parbox[c]{2cm}{\cite{japCue, jap}}} & \makecell[l]{Rhetorical Structure \\Theory\\Linguistic analysis\\Linguistic rules} & \makecell[l]{Claim explanation\\through Description\\segments} & &  \makecell[l]{NTCIR3 data\\59,956 patents} \\
& \\
\parbox[c]{2cm}{\cite{improving}} & \makecell[l]{Discourse-based\\and Deep Syntactic \\Structure-based\\ simplification} & \makecell[l]{Shallow and\\deep strategies} & & 30 patents (test)\\
\parbox[c]{2cm}{\cite{accessible}} & \makecell[l]{Deep Syntactic\\Structure-based\\simplification} & & \makecell[l]{Legal scope\\can be modified\\Complexity} & 500 sentences (test) \\
\parbox[c]{2cm}{\cite{simplificationShallow}} & \makecell[l]{Discourse-structure\\simplification} & & & 29 patents (test) \\
& \\
\parbox[c]{2cm}{\cite{lupu}} & Claim Dependencies\\Graph & \makecell[l]{Adaptation of general\\NLP tools to the\\patent domain} & \makecell[l]{Errors in PoS tagging\\can lead to graph\\collapse} & \makecell[l]{EN CLEF–IP 2012\\Passage Retrieval\\topic set (40 train,\\600 test claims)} \\
& \\
\parbox[c]{2cm}{\cite{ferraro}} & Text segmentation & \makecell[l]{Increase readability \\without modifying\\the text} & \makecell[l]{Body segmentation\\can be improved} & \makecell[l]{821 train, 80 test\\ patents}\\
& \\
\parbox[c]{2cm}{\cite{multiple}} & \makecell[l]{Rules, linguistic\\knowledge, statistics} & \makecell[l]{Text highlighting,\\claims diagram} & \makecell[l]{Complexity\\Linguistic knowledge is\\domain-specific} & 25 patents \\
&\\
\parbox[c]{2cm}{\cite{informationExtraction}} & \makecell[l]{Claim structure analysis\\through Information\\Extraction} & \makecell[l]{Relation extraction\\techniques for \\highlighting\\ Claim aspects} & & \makecell[l]{12,972 patents\\on AI}\\
& \\
\parbox[c]{2cm}{\cite{inria}} & \makecell[l]{Rule-based\\Improve the readability\\of an extracted graph} & \makecell[l]{Machine-oriented \\simplification\\for information\\extraction and\\graph visualization} & \makecell[l]{Simplification does not\\improve extraction\\performance} & 30 patents (test)\\
&\\
\parbox[c]{2cm}{\cite{evalSim}} & User Study & \makecell[l]{Evaluation of users\\attitude toward patents\\and simplification\\solutions} & &\\
\bottomrule
\end{tabular}
}
\caption{Surveyed studies for Patent Simplification.}
\label{tab:simplification}
\end{table}
\cite{ferraro}, for example, aim at improving the Claim's presentation without modifying its text. They segment each claim into preamble, transition, and body (rule-based) and then further divide the body into clauses using a Conditional Random Field. Knowing the elements' boundaries, the claim can then be formatted more clearly, e.g., adding line breaks. 

\begin{figure}
    \centering
    \includegraphics[width=\textwidth]{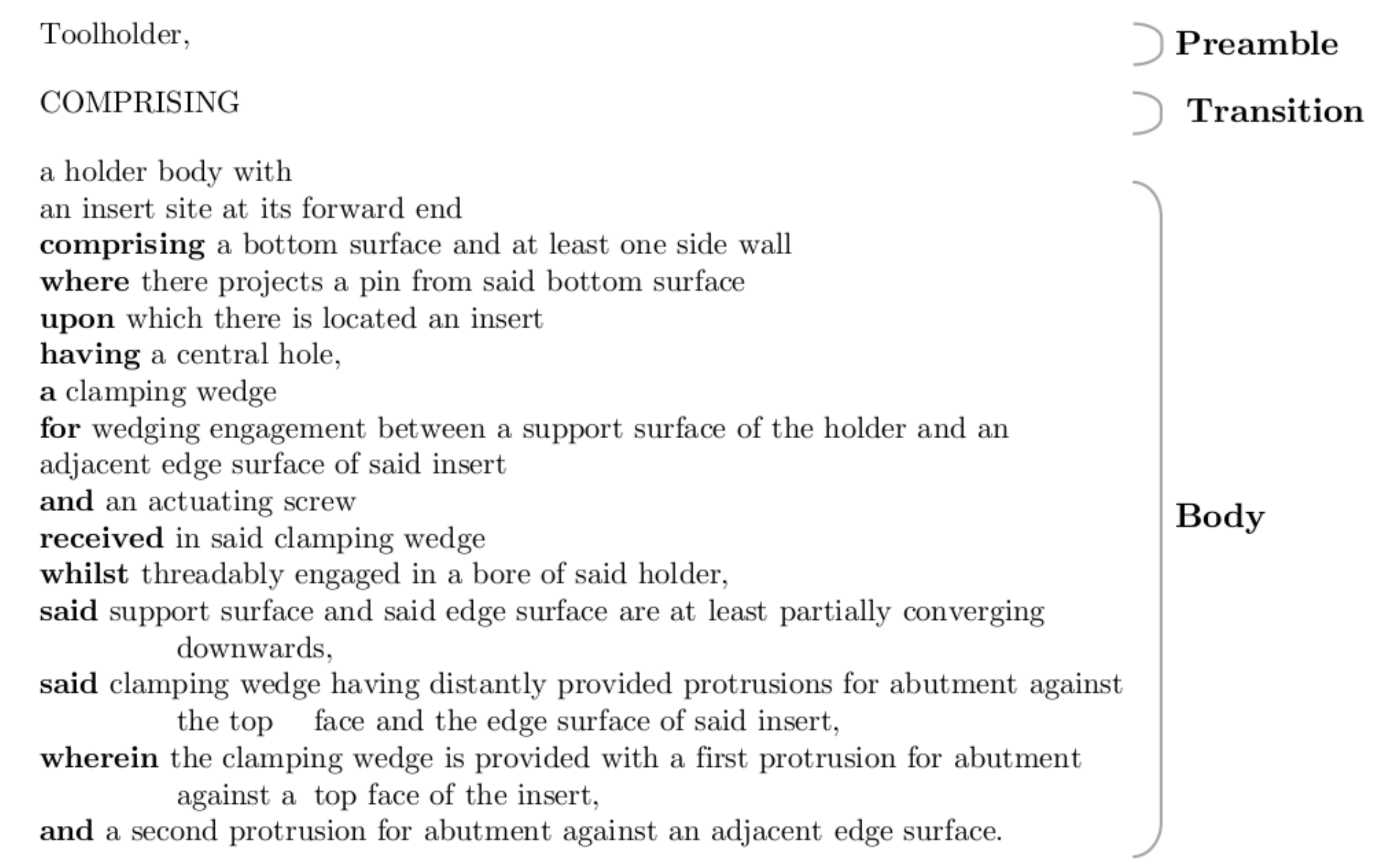}
    \caption{A segmented patent. Adapted from \citep{ferraro}.}
    \label{fig:ferraro}
\end{figure}

A somewhat opposite approach was taken in the PATExpert project~\citep{patexpert}, which developed a rewriting and paraphrasing module~\citep{improving}. The researchers considered two levels of simplification: one uses surface criteria to segments the input and reconstructs chunks into shorter, easier-to-read sentences~\citep{simplificationShallow}. The other~\citep{accessible} is conceptually similar to~\citep{multilingual} for multilingual summarization: after shallow simplification and segmentation, patents are parsed and projected to Deep Syntactic Structures. This representation is in turn used to rewrite a text that is simpler to process for the reader (possibly in another language). Both approaches modify the patent text. Note how, in this framework, rewriting and summarization are essentially unified, with the key difference that no content is removed for simplification. 

Instead of relying on linguistic techniques,~\cite{informationExtraction} use an Information Extraction engine that detects entities types and their relations using distant supervision. They provide a visualization interface which a) formats each patent claims to improve readability: color is used to highlight the claim type (e.g., apparatus, method), the transaction, and technical components in the patent body; b) shows the Claim structure: for each claim they include its type, dependencies, and references to other technologies and components. They target patent experts, which might use the system to compare claims (e.g., in the same patent family) and search for similar documents. 

\begin{figure}
%  \begin{minipage}[b]{0.5\textwidth}
 %   \includegraphics[width=\textwidth]{figures/annotated.JPG}
 % \end{minipage}
  %  \begin{minipage}[b]{0.5\textwidth}
    \includegraphics[width=\textwidth]{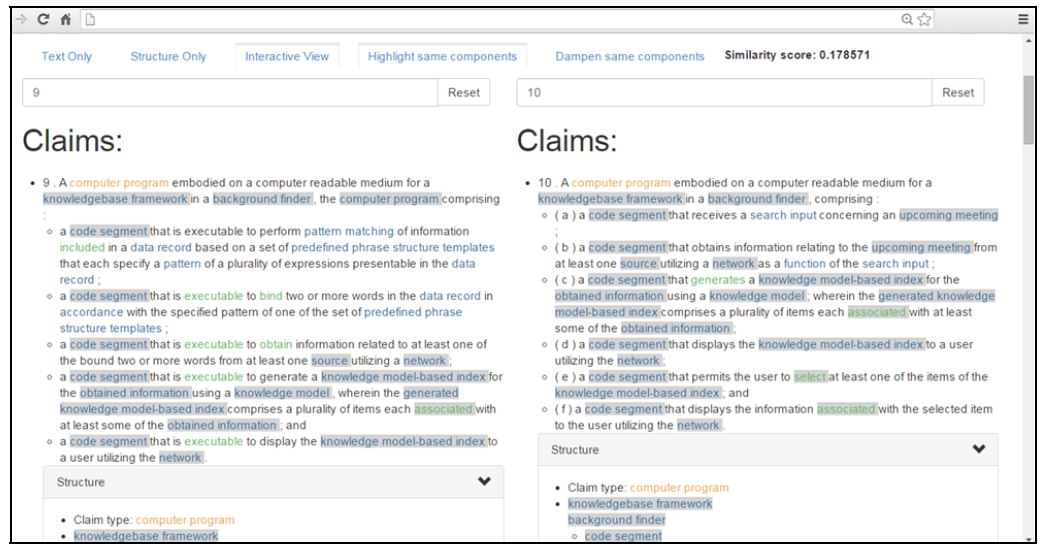}
 % \end{minipage}
  \caption{Interface for comparing two patents, from \citep{informationExtraction}.}
\end{figure}

The approaches described so far output a simplified and easier-to-read textual version of the original Claim. Another option is to visualize them in a structured way. ~\cite{lupu}, for example, obtain a connected graph of the claim content; each node contains a noun phrase (NP) and is linked through a verb, a preposition, or a discourse relation. Similarly,~\citep{inria} constructs a graph for visualizing the patent content in the contest of an Information Retrieval pipeline. \cite{multiple} uses visualization on two levels: they first construct a hierarchical tree of the whole Claim section (highlighting dependency relations) and simplifies each claim. In this phase, a tailored linguistic analysis is used~\citep{nlppatents}; the simplified claim is segmented in shorter phrases (whose NPs are highlighted  and linked to the Description) and visualized as a forest of trees. 

Note that most approaches do not measure the improvement in readability so that it is not clear how effective they are in enhancing intelligibility. 

\begin{figure}
\includegraphics[width=\textwidth]{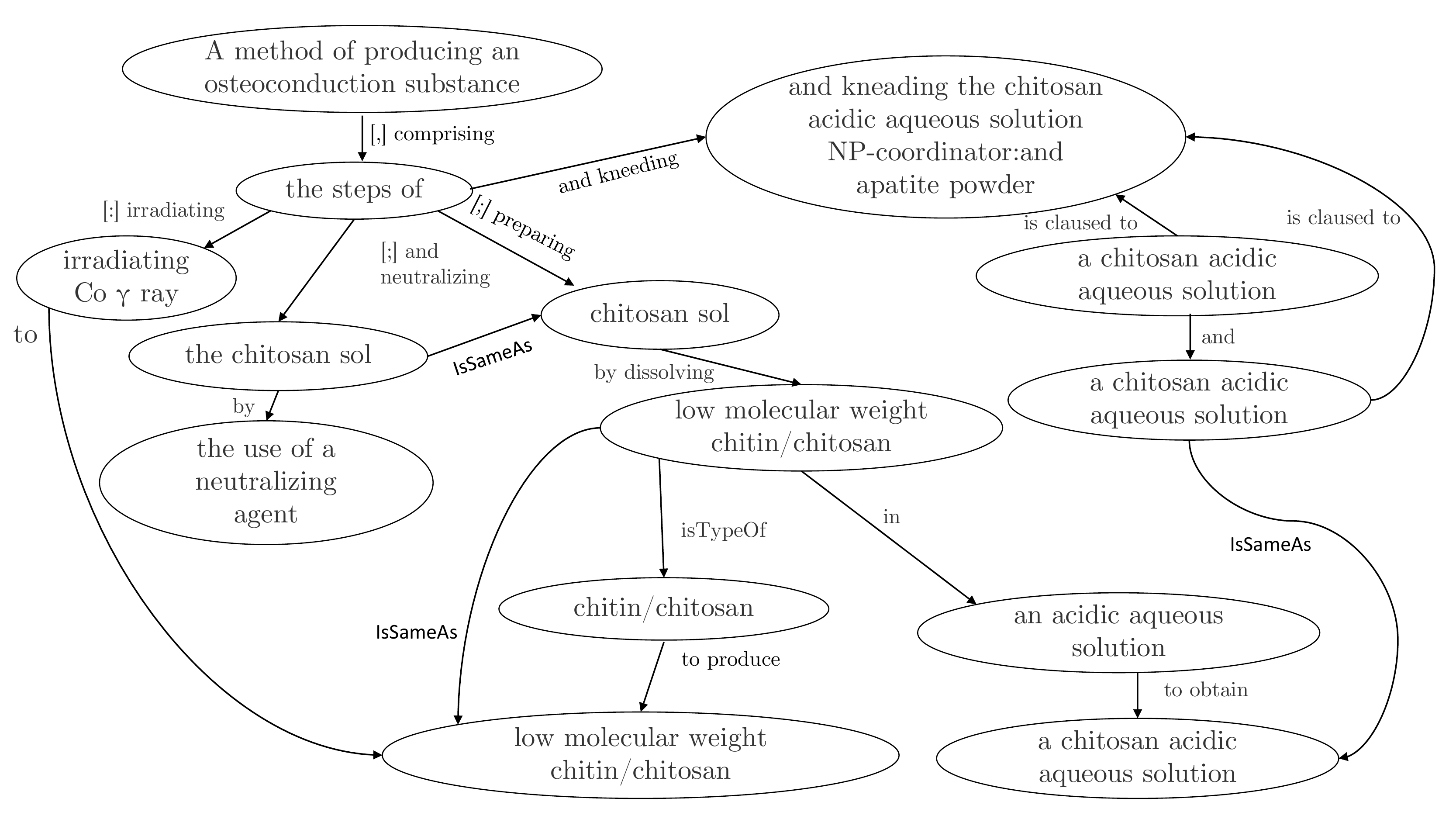}
\includegraphics[width=\textwidth]{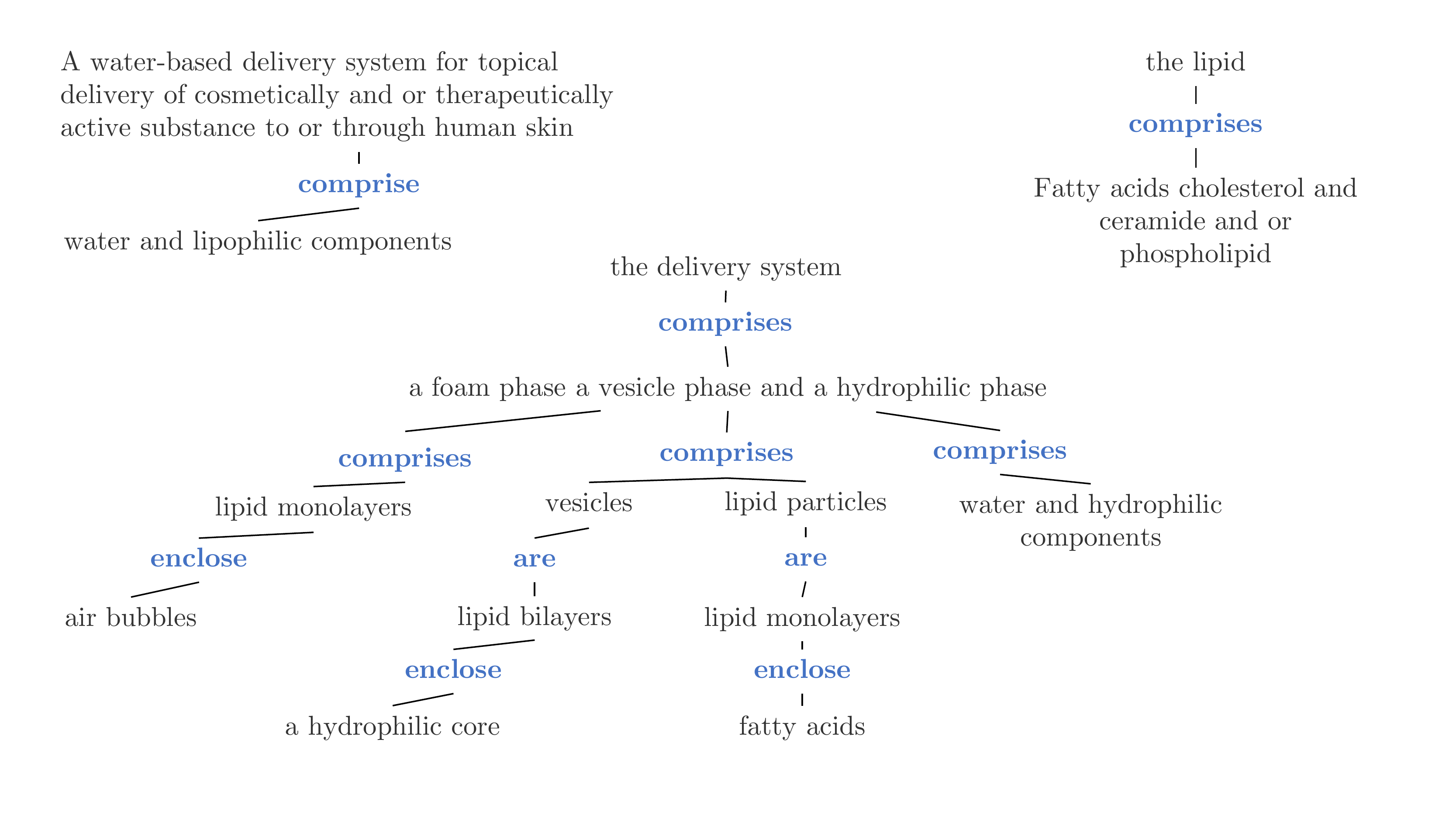}
  \caption{Top: connected graph for visualizing a patent claim, adapted from~\citep{lupu}; bottom: diagram of a claim, adapted from \citep{multiple}.}
\end{figure}

Finally, the Claim simplification problem was also studied for the Japanese language. In particular, Shinmori et al. propose a method to expose patent structure using manually-defined cue phrases~\citep{japCue} and explain invention-specific terms using the Description~\citep{jap}. In~\citep{allignJap}, Description chunks are used to paraphrase corresponding sentences in the Claim and improve readability.   

\section{Approaches for Patent generation}
\label{sec:generation}
The task of Patent generation has recently been investigated by Lee and Hsiang, which try to leverage state-of-the-art NLP models to generate patent text. Table \ref{tab:generation} reports their main results. 

\begin{table}[]
\resizebox{\textwidth}{!}{%
\begin{tabular}{@{}lllll@{}}
\toprule
Study & Approach & Main contribution & Limitations & Dataset \\ \midrule
\parbox[l]{2cm}{\cite{gpt2pat}} & GPT-2 fine-tuning & \makecell[l]{Adaptation of a\\ general-domain LM\\to patent text} & Evaluation & 555,890 patent\\
&\\
\parbox[l]{2cm}{\cite{measuringSpan}} & \makecell[l]{Span-pair classification\\(BERT)} & \makecell[l]{Automatic evaluation\\of generation relevancy} & \makecell[l]{Negative examples\\can have unrelated\\vocabulary} & 14M span pairs\\
&\\
\parbox[l]{2cm}{\cite{metadata}} & \makecell[l]{GPT-2 -based} & \makecell[l]{Conditional generation\\of patent Sections} &  & \makecell[l]{Google Patents\\Datasets\\(1976~2017-08 Utility\\patents)} \\
&\\
\parbox[l]{2cm}{\cite{prior}} & \makecell[l]{Similarity and\\reranking} & \makecell[l]{Ranking of most similar\\training samples to\\the generated text} & Mixed results & Huge \\
\\ \bottomrule
\end{tabular}%
}
\caption{Surveyed studies for Patent Simplification.}
\label{tab:generation}
\end{table}

Their early work~\citep{gpt2pat} fine-tunes GPT-2 -- a language model which demonstrated impressive results in generating text from a wide range of domains -- using patents' first claims. Interestingly, only a small number of fine-tuning steps are sufficient to adapt the general domain model and produce patent-like text. However, the quality of the generation is not measured. This gap is partially filled in~\citep{measuringSpan}, where a BERT classifier is used to measure if two consecutive spans, generated automatically, are consistent. They train the classifier on consecutive spans from the same patent (positive examples) and from non-overlapping classes and subclasses (negative examples), which might make the classification not particularly difficult (e.g., the model could relay in shallow lexical features). The generation process is further investigated in~\citep{prior}, which, given a generated text, tries to find the most similar example in the generator's fine-tuning data.

The models described above try to generate consistent text resembling a patent without specific constrains. \cite{metadata} takes a different route and trains the model to generate a patent's Section (Title, Abstract, or claims) given other parts of the same patents. The model uses GPT-2, which receives as input the text on which to condition and learns to produce a section of the same patent accordingly. For example, one can input the Title of a patent and train the model to generate the corresponding Abstract. Two things should be noted: first, the authors frame the problem as self-supervised and use patents' sections as gold-standard, which simplifies evaluation; second, the problem generalizes abstractive patent summarization, so that it might be interesting to study the performance obtained, e.g., generating the Abstract from the Description. 

\section{Current and future directions}
\label{section:directions}
This survey aimed at showing that patents are an interesting domain both for their practical importance and their linguistic challenges. While generative approaches for patents are still relatively niche topics, with few active groups, the domain is drawing attention from general NLP practitioners for its unique characteristics. 
In the following, we present some open issues which might be worthy of future research.

\begin{table}[]
\resizebox{\textwidth}{!}{%
\begin{tabular}{@{}llll@{}}
\hline
Task & Input $\xrightarrow{}$ Output & Evaluation & Challenges \\ 
\hline
Summarization & \begin{tabular}{@{}l@{}}\makecell[l]{Patent\\or Section}\end{tabular} $\xrightarrow{}$ Summary & \begin{tabular}[l]{@{}l@{}}Human evaluation,\\ROUGE, $F_1$,\\compression,\\retention ratio\end{tabular} &\begin{minipage}[t, topmargin=0pt] {0.25\textwidth} 
      \begin{itemize}[leftmargin=*, topsep=0pt]
      \setlength\itemsep{-0.5em}
      \item Long input
      \item Long sentences
      \item Spread content
      \item Factuality
     \end{itemize} 
    \end{minipage}  \\  
&\\
Simplification & \begin{tabular}[l]{@{}l@{}}Patent text \\ (usually Claim) \end{tabular} $\xrightarrow{}$ \begin{tabular}[c]{@{}l@{}}Simplified text, \\ visual interface\end{tabular} & Human evaluation & \begin{minipage} [t, topsmargin=*] {0.25\textwidth} 
      \begin{itemize}[leftmargin=*,topsep=0pt]
      \setlength\itemsep{-0.5em}
      \item Maintain legal scope
      \item Lack of simplified data
     \end{itemize} 
    \end{minipage} \\
     \\
    
Generation & \begin{tabular}[l]{@{}l@{}}None, \\ seed or Section \end{tabular} $\xrightarrow{}$ Patent text  & \makecell[l]{Human evaluation,\\ROUGE} &\begin{minipage} [t, topsmargin=*]{0.25\textwidth} 
      \begin{itemize}[leftmargin=*,topsep=0pt]
      \setlength\itemsep{-0.5em}
      \item Peculiar language
      \item Domain mismatch
      \item Evaluation
     \end{itemize} 
    \end{minipage} \\
    \\
    \hline
\end{tabular}
}
\caption{The tasks described in this survey and their challenges in the patent domain. In addition, all tasks are challenged by the patents' peculiar linguistic characteristics  described in Section~\ref{section:primer}.}
\label{tab:task}
\end{table}

\begin{description} 
    \item[Data, data, data] Labeled and annotated data are few in the patent domain. For summarization, the only available large-scale dataset is BigPatent~\citep{bigpatent}, while no simplified corpus (let alone parallel corpora) exists, to the best of our knowledge. 
    Moreover, while BigPatent represented a milestone for patent summarization, the target Abstract is written in the typical arcane patent language; thus the practical usefulness of systems trained on these data is probably scarce for laypeople -- which would rather read a ``plain English'' abstract, like those provided by commercial companies. A dataset that targets a clearer summary (unifying summarization and simplification) would also help in understanding models' capabilities in going beyond shallow features and have a global understanding of the source.
    Finally, while no public corpora of simplified patent text exist to date, other domains have exploited creative ploys for minimizing human effort: in the medical domain, for example, \citet{social} uses social media contents to create a simplified corpus. 
    
    \item[Benchmarks] There are many approaches to summarization and simplification. However, it is difficult to compare them given the absence of shared benchmarks. For extractive summarization, for example, many studies have only compared their results with a baseline or a general-domain commercial system. However, directly comparing the performance of different approaches is difficult, as they solve slightly different tasks on different datasets and often fail to report implementation details.  
    \item[Evaluation metrics] Generative approaches for patent often resort to general-domain metrics for evaluation (e.g. ROUGE). However, it is not clear how suitable these measures are for the patent domain, given its peculiarities. In the context of abstractive summarization and patent generation, some works~\citep{namingabstractive, metadata} highlight that ROUGE is unable to find semantically similar sentences expressed in different wording. 
    In the context of Natural Language Generation, some new measures have recently been proposed to solve these issues. BERTScore~\citep{bertscore}, for example, evaluates the similarity among the summary and gold-standard tokens instead of their exact match, while QAGS~\citep{qaeval} uses a set of questions to evaluate factual consistency between a summary and its source (a reference is not needed). It is yet to be explored if these metrics could be applied to the patent domain successfully. Finally, note  that even human studies are difficult in the patent domain, as they require a high expertise, which most people lack.
    \item[Factuality] While neural abstractive models have shown impressive performance in summarization, they tend to fabricate information. \cite{factful} studied the phenomenon in the news domain and found that around 30\% of documents included fake facts. This behavior is particularly problematic in a legal context; ROUGE, however, is a surface metric and is unable to detect factual inconsistencies.
    \item[Domain adaptation] Patents' language  hardly resembles general-discourse English (used in pre-training), but the domain adaptation problem has not been studied in detail. Among the previous works,~\cite{multidomain} propose a second multitask pre-training step,~\cite{domain} studies models cross domain performance and~\cite{few} evaluates zero and few shot settings; all these works described applications to the patent domain, among the others. 
    \item[Input length] Patent documents are extremely long. For summarization, the only datasets which have comparable or longer inputs are the arXiv and the PubMed dataset~\citep{arxiv_pubmed}, which summarize entire research papers.  While solutions to allow the processing of long inputs have been proposed, the in-depth study of methods and performance for such long documents is still in its early days. For neural models, a very long input translates into prohibitive computational requirements (e.g. several GPUs), which researchers have recently tried to mitigate by modifying the underlying architectures.
\end{description}

\bibliography{sample}

\end{document}